\documentclass{article}

\usepackage[preprint]{_sty/neurips_2026}

\usepackage[utf8]{inputenc} 
\usepackage[T1]{fontenc}    
\usepackage{hyperref}       
\usepackage{url}            
\usepackage{graphicx}       
\usepackage{booktabs}       
\usepackage{multirow}       
\usepackage{amsfonts}       
\usepackage{nicefrac}       
\usepackage{microtype}      
\usepackage[table]{xcolor}  
\usepackage{amsmath}  
\usepackage{amssymb}
\usepackage{capt-of}
\usepackage{wrapfig}
\usepackage{subcaption}
\usepackage{placeins}

\newbool{inccomment}
\booltrue{inccomment}  

\definecolor{mygray}{RGB}{230, 230, 230}

\newcommand{\XC}[1]{\ifbool{inccomment}{{\color{magenta}XC\@: #1}}{}}
\newcommand{\JY}[1]{\ifbool{inccomment}{{\color{blue}JY\@: #1}}{}}

\title{Pyramid Forcing: Head-Aware Pyramid KV Cache Policy for High-Quality Long Video Generation}

\author{
Jiayu Chen\textsuperscript{1}
\quad
Junbei Tang\textsuperscript{2}\thanks{Work performed during an internship at Peking University.}
\quad
Wenbiao Zhao\textsuperscript{3}\footnotemark[1]
\quad
Maoliang Li\textsuperscript{1}
\\
\textbf{Jiayi Luo\textsuperscript{4,5}}
\quad
\textbf{Zihao Zheng\textsuperscript{1}}
\quad
\textbf{Jiawei Yang\textsuperscript{1}}
\quad
\textbf{Guojie Luo\textsuperscript{1}}
\quad
\textbf{Xiang Chen\textsuperscript{1}\thanks{Corresponding author.}}
\\
\textsuperscript{1}Peking University
\quad
\textsuperscript{2}South China University of Technology
\quad
\textsuperscript{3}Xinjiang University
\\
\textsuperscript{4}Beihang University
\quad
\textsuperscript{5}Zhongguancun Academy
}

\begin{document}

\maketitle
\begin{center}
    \includegraphics[width=\textwidth]{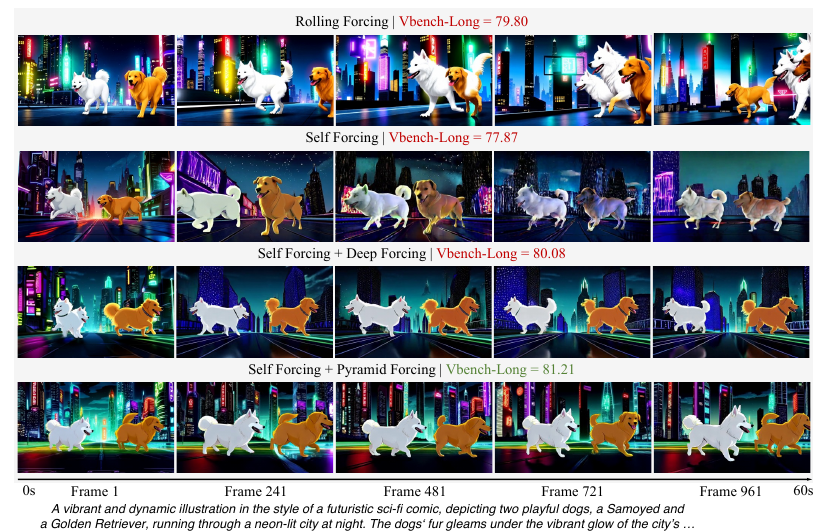}
    \captionof{figure}{
    Pyramid Forcing mitigates long-video degradation, including appearance drift and subject inconsistency, achieving a state-of-the-art VBench-Long score of 81.21 among all baselines.
    }
    \label{fig:1}
\end{center}

\begin{abstract}
    Autoregressive video generation enables streaming and open-ended long video synthesis, but still suffers from long-term degradation caused by accumulated errors. Existing KVCache strategies usually apply unified historical-frame retention, implicitly assuming homogeneous historical dependencies across attention heads. We revisit historical-frame attention and reveal three distinct head types: Anchor Heads require broad long-range context, Wave Heads exhibit periodic temporal dependencies, and Veil Heads focus on initial and adjacent frames. Based on this finding, we propose \textbf{Pyramid Forcing}, a head-aware pyramidal KVCache framework that identifies head types offline, assigns behavior-specific cache policies, and supports heterogeneous cache lengths via efficient ragged-cache attention. Experiments on Self Forcing and Causal Forcing show that Pyramid Forcing consistently improves long-horizon generation quality on VBench-Long, increasing the 60-second Self Forcing score from 77.87 to 81.21 while enhancing motion dynamics, visual fidelity, and semantic consistency. Project: \url{https://if-lab-pku.github.io/Pyramid-Forcing/}.
\end{abstract}

\begin{figure*}
    \includegraphics[width=\textwidth]{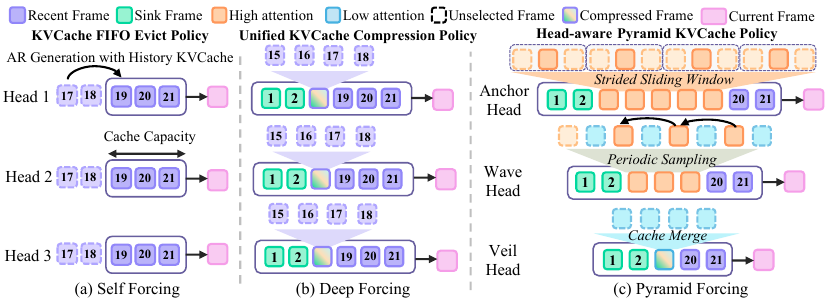}
    \caption{
    Comparison of KVCache policies. Unlike Self Forcing and Deep Forcing with unified cache compression, Pyramid Forcing assigns head-aware strategies to Anchor, Wave, and Veil Heads for differentiated historical information allocation.
    }
    \label{fig:2}
\end{figure*}

\section{Introduction}
\label{sec:intr}

Autoregressive (AR) video generation models \citep{huang2025self,teng2025magi,zhu2026causal} provide a natural pathway for long video generation by iteratively denoising future frames conditioned on previously generated frames.
    Unlike diffusion-based video generation models \citep{peebles2023scalable,wan2025wan,yang2024cogvideox} constrained by a fixed temporal scope, AR video generation supports streaming generation, continuation and open-ended generation, making it promising for applications such as interactive generation\citep{he2025matrix,mao2025yume}, world simulation\citep{yuan2026inference,xie2026generated}, and embodied intelligence\citep{feng2025vidarc,ye2026world}.
However, in practical scenarios, long video generation still suffers from long-term degradation. As illustrated in Fig. ~\ref{fig:1}, as the generation process progresses, visual appearance tends to drift, motion dynamics weaken, and object identity gradually becomes blurred.

Existing works~\citep{cui2025self} have suggested that this degradation primarily arises from error accumulation during autoregressive generation, especially when full historical information is retained.
    To address this issue, prior methods attempt to mitigate error accumulation by retaining recent frames\citep{yang2025longlive,liu2025rolling}, preserving early target frames\citep{li2026rolling,kim2026memrope}, or selecting important historical information\cite{yi2025deep}.
However, these approaches remain limited in alleviating error accumulation. As shown in Fig.~\ref{fig:2}, we further observe that most of them implicitly assume a uniform dependency pattern on historical information within the model, thereby adopting a unified coarse-grained cache compression strategy.

To examine this assumption, we revisit the attention mechanism in autoregressive video generation models represented by Self Forcing, and analyze the dependency patterns and temporal behaviors of different attention heads with respect to historical information during long-term generation.

Our analysis reveals significant heterogeneity in how attention heads utilize historical information:
    One group of heads maintains strong and stable attention over a wide range of historical frames, exhibiting long-range anchoring capability, which we term \textbf{Anchor Heads}.
    Another group shows periodic temporal fluctuations in attention over historical frames, demonstrating dynamic and oscillatory temporal preferences, which we term \textbf{Wave Heads}.
    A small subset of heads exhibits weak attention to recent history but is highly sensitive to the initial frame, where excessive historical input can even suppress their focus on the first frame, which we term \textbf{Veil Heads}.

These findings indicate that the implicit assumption of uniform compression in existing methods does not hold. Different attention heads exhibit distinct patterns of dependency on historical information. Therefore, historical information should be allocated in a differentiated manner based on head-specific temporal behaviors to more effectively preserve critical context and suppress error accumulation.

Based on this observation, we propose Pyramid Forcing, a head-aware pyramidal KVCache framework for long video generation, as shown in Fig.~\ref{fig:2}(c). 
    Specifically, we first introduce \textbf{Offline Tri-Pattern Head Classification}, which uses sign-rate statistics and frequency-domain periodicity to identify Anchor, Wave, and Veil Heads from a small calibration set. 
    Then, we introduce head-specific \textbf{Pyramid KVCache Policies} tailored to distinct historical-frame attention patterns, using strided retention for Anchor Heads, period-aligned sampling for Wave Heads, and compact local merging for Veil Heads to balance long-range coverage, periodic dependency, and local continuity.
    We further introduce \textbf{Heterogeneous KVCache Computation Optimization}, enabling low-overhead ragged-cache attention for variable head-level cache lengths.

We evaluate Pyramid Forcing on Self Forcing and Causal Forcing using VBench-Long over 30s and 60s generations. Pyramid Forcing consistently improves long-horizon quality across both backbones, increasing the 60s total score of Self Forcing from 77.87 to 81.21. It also yields substantial gains in dynamic degree, with improvements of up to 29.36 points on Causal Forcing, demonstrating stronger motion dynamics, visual fidelity, and semantic consistency in long video generation.



\section{Preliminaries}

\subsection{Autoregressive Video Diffusion}

Autoregressive video diffusion models generate videos frame by frame during the denoising diffusion process, conditioned on previously generated frames. Given a video sequence of $N$ frames, its joint probability distribution can be factorized in an autoregressive manner as $p(x^{1:N}) = \prod_{t=1}^{N} p(x^t \mid x^{<t})$.
    Specifically, the previously generated frames $x^{<t} = (x^1, x^2, \ldots, x^{t-1})$ are modeled by the self-attention mechanism and used to predict the current frame $x^t$.

The query of the current frame, $Q \in \mathbb{R}^{B \times H \times L_q \times D}$, attends to the keys and values of historical frames, $K, V \in \mathbb{R}^{B \times H \times L_{kv} \times D}$. The attention operation is computed as
\begin{equation}
\operatorname{Attn}(Q, K, V)
=
\operatorname{softmax}
\left(
\frac{QK^\top}{\sqrt{D}}
\right)V .
\end{equation}
Here, $H$ denotes the number of attention heads, and $D = C/H$ is the channel dimension of each head. $L_q$ and $L_{kv}$ denote the sequence lengths of the query and key/value tokens, respectively. The resulting attention logits typically have shape
$QK^\top \in \mathbb{R}^{B \times H \times L_q \times L_{kv}}.$

\subsection{KVCache Policy in Autoregressive Video Diffusion}

During autoregressive video generation, historical frames are encoded into latent tokens and stored as KVCaches. To reduce computation and mitigate error accumulation, existing methods maintain a rolling cache by feeding only selected historical frames into the model. When generating the $t$-th frame, a subset $\mathcal{I}(t) \subset \mathcal{H}(t)$ is selected from the candidate history, typically consisting of sink frames $\mathcal{S}(t)$ for global anchoring and recent frames $\mathcal{R}(t)$ for local temporal context. Their KVCaches are concatenated as
$K' = [K_{\text{sink}} \mid K_{\text{recent}}],
\quad
V' = [V_{\text{sink}} \mid V_{\text{recent}}],$
and attention is computed by
\begin{equation}
\operatorname{Attn}(Q_t, K'_t, V'_t)
=
\operatorname{softmax}
\left(
\frac{
Q_t [K_{\text{sink}}^\top, K_{\text{recent}}^\top]
}{\sqrt{D}}
\right)
[V_{\text{sink}}, V_{\text{recent}}].
\end{equation}
After each generation step, the new keys and values are written into $\mathcal{H}(t+1)$ and used to update $\mathcal{I}(t+1)$. Existing methods usually apply a unified cache policy across attention heads; in contrast, we analyze head-specific historical dependencies and optimize KV caching at a finer granularity.
\begin{figure*}
    \includegraphics[width=\textwidth]{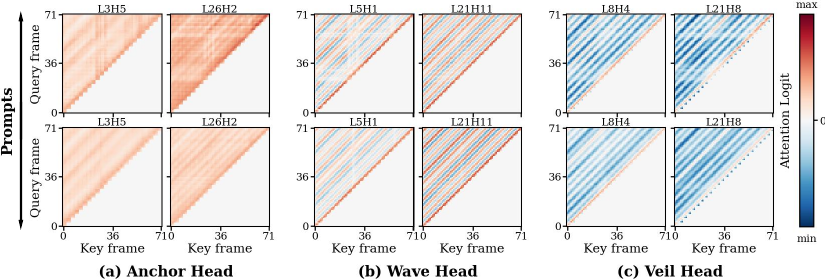}
    \caption{
Historical-frame attention patterns of three head types in Self Forcing. 
\textit{Anchor Heads} capture broad long-range dependencies, \textit{Wave Heads} exhibit periodic temporal selectivity, and \textit{Veil Heads} focus on the first and adjacent frames while suppressing mid-history.
    }
    \label{fig:3}
\end{figure*}

\begin{figure}
    \centering
    \includegraphics[width=\linewidth]{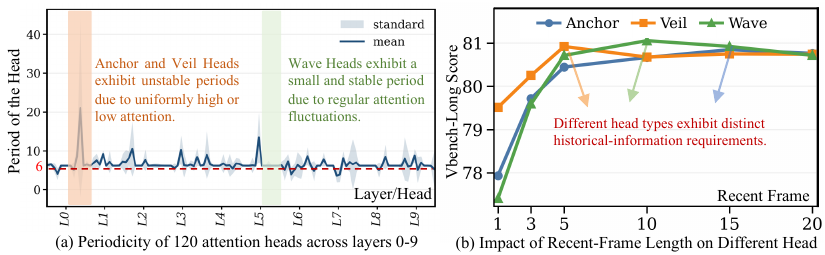}
    \caption{
Analysis of head-wise periodicity and historical information demands.
(a) Wave Heads exhibit a small and stable fluctuation period under FFT analysis, whereas others lack stable periodicity.
(b) Different head types prefer distinct recent history, motivating a head-aware KVCache policy.
}
    \label{fig:4}
\end{figure}

\section{Motivation}

Autoregressive video generation predicts future frames by attending to the KVCache of historical frames. However, irrelevant historical frames may introduce noise and exacerbate error accumulation, leading to pronounced long-term degradation and temporal drift in extended generation. To better inform KVCache policy design, we empirically analyze historical-frame attention patterns in Self Forcing and summarize the observations as follows.

\paragraph{Observation 1: Attention heads exhibit significant heterogeneity in their historical-frame attention patterns.}
\label{obser:1}
Based on the pre-softmax attention logits and their corresponding historical-frame attention patterns, we categorize attention heads into three types: \emph{Anchor Heads}, \emph{Wave Heads}, and \emph{Veil Heads}, as illustrated in Fig.~\ref{fig:3}.

\textit{Anchor Heads.}
As shown in Fig.~\ref{fig:3}(a), Anchor Heads assign high attention scores across broad historical ranges, forming globally activated attention maps. This pattern suggests a long-range anchoring role that preserves historical information and supports temporal consistency. Accordingly, Anchor Heads require cache policies that retain wider temporal coverage under limited cache budgets.

\textit{Wave Heads.}
As shown in Fig.~\ref{fig:3}(b), Wave Heads exhibit regular positive--negative alternations along the temporal dimension, forming wave-like attention maps. This pattern indicates strong temporal fluctuations and selective attention to intermediate historical frames, which helps preserve temporal consistency in long video generation. Thus, Wave Heads only need to retain temporally patterned key frames, reducing errors from non-critical history.

\textit{Veil Heads.}
As shown in Fig.~\ref{fig:3}(c), Veil Heads concentrate attention on the first and adjacent frames, while assigning negative scores to most intermediate frames, yielding globally low attention-score maps. This pattern suggests reliance on initial and local neighboring frames for structural information and core semantics, with little mid-history dependence. Thus, retaining excessive intermediate frames may suppress key signals, introduce errors, and degrade generation quality.

Moreover, despite variations in input prompts, the three types of attention heads consistently exhibit stable attention patterns. This consistency suggests that attention heads in the autoregressive video generation model possess stable functional roles that are intrinsic to the model, making them reliably classifiable. More attention visualization results are provided in the supplementary material.

\paragraph{Observation 2: Wave Heads exhibit periodic attention patterns over historical frames.}
\label{obser:2}
To characterize the temporal fluctuations of Wave Heads, we apply FFT to the attention weights from the last generated frame to all historical frames in Self Forcing. As shown in Fig.~\ref{fig:4}(a), Wave Heads exhibit a small and stable fluctuation period, unlike Anchor and Veil Heads. This suggests that, when generating the $t$-th frame, Wave Heads consistently attend to frames at positions $t - P \times i,\ i \geq 1$, where $P$ denotes the estimated period. Therefore, Wave Heads can be identified by periodicity, and their key historical frames can be retained accordingly. See Appendix ~\ref{ap:wave} for theoretical analysis.

\paragraph{Observation 3: Different attention heads require different amounts of historical information.}
\label{obser:3}
As shown in Fig~\ref{fig:4}(b), we evaluate how the length of recent frame $\mathcal{R}(t)$ affects 30-second video generation quality across the three head types. Anchor Heads improve consistently with longer windows, confirming their dependence on long-range history. Wave Heads perform best with 10 retained frames, and Veil Heads peak at only 5 frames, with quality degrading as the window further expands. These results reveal distinct head-wise demands for historical information, suggesting that AR video generation should adopt head-aware KVCache allocation instead of the unified policies.
\begin{figure*}
     \centering
    \includegraphics[width=\textwidth]{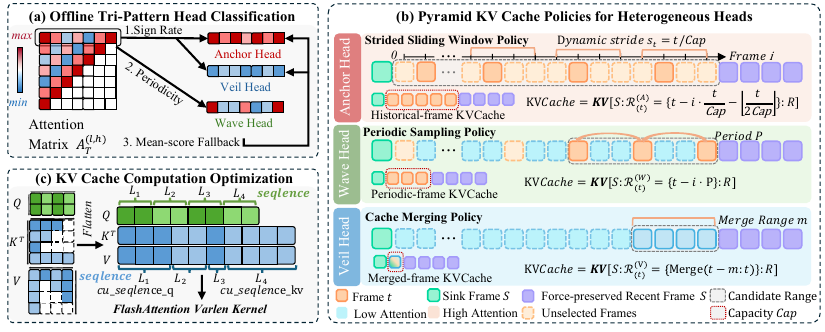}
    \caption{
    Overview of Pyramid Forcing.
    (a) Offline Tri-Pattern Head Classification identifies Anchor, Wave, and Veil Heads.
    (b) Pyramid KVCache Policies assign head-specific cache strategies.
    (c) KVCache Computation Optimization enables efficient ragged-cache attention with optimized kernels.
    }
    \label{fig:5}
\end{figure*}

\section{Method}

\subsection{Offline Tri-Pattern Head Classification}

Motivated by the heterogeneous yet stable historical attention distributions in video generation, we propose an offline head classification method(Fig.~\ref{fig:5}(a)). We first sample a small set of prompts, generate videos, and classify attention heads according to their historical-frame attention patterns.

As discussed in \textbf{Observation~1} and \textbf{Observation~2}, Anchor Heads assign broadly positive scores to historical frames, Veil Heads mainly attend to the first and nearby frames while suppressing most history, and Wave Heads exhibit periodic positive--negative oscillations. We therefore classify heads using two complementary criteria: sign-rate statistics and frequency-domain periodicity.

Specifically, when generating the $T$-th frame, for the $h$-th head in the $l$-th layer, we denote the pre-softmax attention-score matrix as $A_T^{(l,h)} \in \mathbb{R}^{T \times T}$ and extract the historical attention sequence as $a^{(l,h)}(t)=A_T^{(l,h)}[T,t]$, $t=1,\ldots,T-1$. We then compute the sign rates of  sequence $a^{(l,h)}(t)$ as
\begin{equation}
r_{\mathrm{pos}}
=
\frac{1}{T-1}
\sum_{t=1}^{T-1}
\mathbf{1}\{a^{(l,h)}(t)>0\},
\quad
r_{\mathrm{neg}}=1-r_{\mathrm{pos}}.
\end{equation}
Given a threshold $\alpha$ adjusted according to the empirical attention-score distribution, heads with $r_{\mathrm{pos}}\geq\alpha$ are classified as Anchor Heads, while those with $r_{\mathrm{neg}}\geq\alpha$ are classified as Veil Heads.

For the remaining heads, we detect periodicity in the frequency domain. After preprocessing $a^{(l,h)}(t)$, we apply FFT and estimate the dominant period $P=\operatorname{Period}(\operatorname{FFT}(a^{(l,h)}(t)))$. Since Wave Heads concentrate within a small and stable period range, a head is classified as a Wave Head if $P<\beta$, where $\beta$ is determined by theoretical and experimental analysis.

For any still-unclassified heads, we use a mean-score fallback: $\bar{a}^{(l,h)}=\frac{1}{T-1}\sum_{t=1}^{T-1}a^{(l,h)}(t)$. If $\bar{a}^{(l,h)}>0$, the head is assigned to Anchor Heads; otherwise, it is assigned to Veil Heads. Finally, we obtain the layer-wise head sets $\mathcal{H}_A$, $\mathcal{H}_W$, and $\mathcal{H}_V$, which are mutually exclusive and jointly cover all attention heads. For notational simplicity, we use superscripts $(A)$, $(W)$, and $(V)$ to denote variables associated with the three head types.

\subsection{Pyramid KVCache Policies for Heterogeneous Heads}

As shown in Fig.~\ref{fig:5}(b), we introduce Pyramid KVCache Policies, which adapt historical-frame caching to Anchor, Wave, and Veil Heads, enabling fine-grained head-aware allocation for inference.

\paragraph{Anchor Heads KVCache Policy.}
Anchor Heads attend to a broad range of historical frames, requiring a cache policy that preserves long-range information under a limited budget. We therefore adopt an \emph{Adaptive Strided Sliding Window} strategy, which adjusts the stride according to the current history length and expands the effective coverage to the full generated sequence.

Specifically, when generating the $t$-th frame, we set the dynamic stride as $s_t = t / \mathrm{Cap}$ and uniformly sample $\mathrm{Cap}$ representative historical frames by selecting the center of each strided window:
\begin{equation}
\mathcal{R}^{(A)}_{(t)}
=
\left\{
t - i \cdot \frac{t}{\mathrm{Cap}}
- \left\lfloor \frac{t}{2\mathrm{Cap}} \right\rfloor
\mid
i = 0,1,\ldots,\mathrm{Cap}-1
\right\}.
\end{equation}
The corresponding KVCache is then
$\mathbf{K}_{\mathrm{recent}}^{(A)}
=
\mathbf{K}\big[\mathcal{R}^{(A)}_{(t)},:\big],
\quad
\mathbf{V}_{\mathrm{recent}}^{(A)}
=
\mathbf{V}\big[\mathcal{R}^{(A)}_{(t)},:\big].$
This centered strided sampling preserves broad historical semantics in a sparse yet representative manner.

\paragraph{Wave Heads KVCache Policy.}
Wave Heads exhibit stable periodic attention over historical frames, requiring cache retention aligned with their intrinsic oscillation. We therefore adopt a \emph{Periodic Sampling} policy, which selects representative historical frames across multiple periods under a limited budget.

Specifically, when generating the $t$-th frame, we use the dominant period $P$ identified by the offline analysis in Sec.~\ref{obser:2} and retain $\mathrm{Cap}$ historical frames at interval $P$:
\begin{equation}
\mathcal{R}^{(W)}_{(t)}
=
\left\{
t - i \cdot P
\mid
i = 0,1,\ldots,\mathrm{Cap}-1
\right\}.
\end{equation}
The corresponding KVCache is then 
$\mathbf{K}_{\mathrm{recent}}^{(W)}
=
\mathbf{K}\big[\mathcal{R}^{(W)}_{(t)},:\big],
\quad
\mathbf{V}_{\mathrm{recent}}^{(W)}
=
\mathbf{V}\big[\mathcal{R}^{(W)}_{(t)},:\big].$
This period-aligned sampling captures the fluctuating long-range dependency of Wave Heads.

\paragraph{Veil Heads KVCache Policy.}
Veil Heads concentrate attention on the first and nearby frames, while assigning negative scores to intermediate history. To avoid errors from excessive historical context, we adopt a \emph{Cache Merging} strategy that compresses local information within a small window.

Specifically, when generating the $t$-th frame, we split each frame in a local window of size $m$ into $m$ feature subsets. We then take the corresponding $1/m$ subset from each frame and concatenate them into a representative feature:
$\mathcal{R}^{(V)}_{(t)}
=
\left\{
\operatorname{merged}(t-m:t)
\right\}.$
The merged KVCache is defined as
$\mathbf{K}_{\mathrm{merged}}^{(V)}
=
\operatorname{Concat}_{i=0}^{m-1}
\mathbf{K}\big[t-m+i,\mathcal{C}_i\big],
\quad
\mathbf{V}_{\mathrm{merged}}^{(V)}
=
\operatorname{Concat}_{i=0}^{m-1}
\mathbf{V}\big[t-m+i,\mathcal{C}_i\big],$
where $\mathcal{C}_i$ denotes the selected feature subset of the $i$-th frame. This local merging preserves core nearby information while suppressing irrelevant intermediate history.

In addition, all heads share the same sink frames $S$ and forced recent frames $R$, as both receive high attention across head types. We further apply Dynamic RoPE to sink frames following prior works~\cite{yi2025deep,yesiltepe2025infinity}. Pyramid Forcing differs by applying head-aware policies to intermediate historical frames, allocating retention ranges and cache strategies according to heterogeneous temporal dependencies.

\subsection{Heterogeneous KVCache Computation Optimization}

\paragraph{Ragged-Cache Attention. }
In standard autoregressive video diffusion, attention assumes a shared dense KVCache, $K,V\in\mathbb{R}^{B\times H\times L\times D}$. Our head-specific cache policies break this uniform-length assumption by assigning each batch--head pair an effective history length $L_{\langle b,h\rangle}$, yielding a head-level ragged cache. Direct dense attention is thus inapplicable, while padding incurs redundant memory and computation.
    To support heterogeneous head-level caches, we propose Ragged-Cache Attention, which stores variable-length KVCaches as flattened sequences with boundary indices, as shown in Fig.~\ref{fig:5}(c). Specifically, given $N=B\times H$ sequences with lengths $L_i$, we define $\mathrm{cu\_seqlens}_i=\sum_{j=0}^{i-1}L_j$ for $1\leq i\leq N$, such that the $i$-th sequence occupies $[\mathrm{cu\_seqlens}_i,\mathrm{cu\_seqlens}_{i+1})$. With queries and KVCaches flattened in batch--head order, variable-length FlashAttention\citep{dao2022flashattention} computes attention directly from flattened $Q$, $K/V$, and $\mathrm{cu\_seqlens}$ without padding.

\paragraph{Efficient kernel customization.}
Although variable-length FlashAttention\citep{dao2022flashattention} supports packed ragged sequences, separate per-frame calls introduce nontrivial launch overhead. We therefore customize execution with multi-frame attention fusion, which merges three per-layer FlashAttention-varlen calls into one ragged-segment invocation after permuting queries to match the KV layout. We further vectorize workspace packing by replacing many per-head slice writes with batched writes. These optimizations reduce attention-kernel launches by more than two-thirds and lower GPU scheduling calls from about 60 to 6, enabling efficient heterogeneous KVCache computation.

\section{Experiments}

\begin{table*}[t]
\small
\centering
\setlength{\tabcolsep}{3pt}
\renewcommand{\arraystretch}{1.15}
\caption{Quantitative comparison on long video generation. We evaluate Pyramid Forcing against autoregressive video diffusion generation baselines for 30- and 60-second videos using VBench-Long.}
\vspace{4pt}
\label{tab:main_results}
\resizebox{\textwidth}{!}{%
\begin{tabular}{lcccccccc}
\toprule
\multicolumn{1}{c}{\multirow{2}{*}{Model}} &
Dynamic & Motion & Overall & Imaging & Aesthetic & Quality & Semantic & Total \\
& Degree $\uparrow$ & Smoothness $\uparrow$ & Consistency $\uparrow$ & Quality $\uparrow$ & Quality $\uparrow$ & Score $\uparrow$ & Score $\uparrow$ & Score $\uparrow$ \\
\midrule
& \multicolumn{8}{c}{\textit{30 seconds}} \\
\cmidrule{2-9}
CausVid             & 45.70 & 98.25 & 22.39 & 66.11 & 59.63 & 86.66 & 50.93 & 79.51 \\
Rolling Forcing     & 30.91 & 98.72 & 24.85 & 71.43 & 61.69 & 86.18 & 55.20 & 79.98 \\
LongLive            & 41.70 & 98.79 & 24.54 & 68.79 & 61.36 & 86.97 & 54.31 & 80.44 \\
\cmidrule{2-9}
Self Forcing        & 44.34 & 98.52 & 24.66 & 70.66 & 63.22 & 87.14 & 54.31 & 80.57 \\
+ Deep Forcing      & 46.48 & 98.43 & 25.10 & 71.70 & 63.77 & 87.23 & 54.00 & 80.59 \\
\rowcolor{gray!12}
+ Pyramid Forcing   & \textbf{55.07} & \textbf{98.82} & \textbf{25.26} & \textbf{72.17} & \textbf{66.55} & \textbf{88.93} & \textbf{55.52} & \textbf{82.25} \\
\cmidrule{2-9}
Causal Forcing      & 61.38 & \textbf{98.48} & 23.03 & \textbf{69.66} & 58.22 & 86.45 & 53.12 & 79.78 \\
\rowcolor{gray!12}
+ Pyramid Forcing   & \textbf{82.67} & 97.23 & \textbf{23.21} & 68.64 & \textbf{60.03} & \textbf{86.98} & \textbf{54.07} & \textbf{80.40} \\
\midrule
& \multicolumn{8}{c}{\textit{60 seconds}} \\
\cmidrule{2-9}
CausVid             & 43.67 & 98.28 & 21.65 & 65.54 & 59.34 & 86.58 & 49.87 & 79.23 \\
Rolling Forcing     & 32.08 & 98.73 & 24.20 & 70.55 & 61.06 & 86.19 & 54.20 & 79.80 \\
LongLive            & 40.93 & 98.77 & 24.54 & 68.83 & 61.70 & 86.89 & 54.79 & 80.47 \\
\cmidrule{2-9}
Self Forcing        & 43.75 & 97.85 & 22.33 & 64.28 & 55.62 & 84.84 & 50.01 & 77.87 \\
+ Deep Forcing      & 43.10 & 98.35 & 24.48 & 68.43 & 60.48 & 86.37 & 54.91 & 80.08 \\
\rowcolor{gray!12}
+ Pyramid Forcing   & \textbf{53.68} & \textbf{98.53} & \textbf{24.70} & \textbf{70.03} & \textbf{62.00} & \textbf{87.58} & \textbf{55.71} & \textbf{81.21} \\
\cmidrule{2-9}
Causal Forcing      & 57.03 & \textbf{98.49} & \textbf{22.51} & \textbf{68.63} & 56.87 & 85.84 & 52.36 & 79.14 \\
\rowcolor{gray!12}
+ Pyramid Forcing   & \textbf{86.39} & 97.03 & 22.41 & 68.38 & \textbf{58.91} & \textbf{86.64} & \textbf{53.05} & \textbf{79.92} \\
\bottomrule
\end{tabular}%
}
\end{table*}

\subsection{Experimental Setup}

\paragraph{Evaluation.}
Following Self Forcing~\citep{huang2025self}, we use 128 MovieGen~\citep{polyak2024movie} prompts for 30-second and 60-second video generation. All results are evaluated on VBench-Long~\citep{huang2024vbench,huang2025vbench++}, with representative metrics reported in the main paper and full results provided in the appendix~\ref{sec:appendix_main}. We compare Pyramid Forcing with autoregressive video diffusion baselines, including distillation-based methods such as CausVid~\citep{yin2025slow}, Rolling Forcing~\citep{liu2025rolling}, and LongLive~\citep{yang2025longlive}, as well as the training-free Deep Forcing~\citep{yi2025deep}. Unless otherwise specified, all methods share the same inference settings, sampling hyperparameters, and generation lengths, differing only in KVCache mechanisms.

\paragraph{Implementation Details.}
We implement Pyramid Forcing on Self Forcing~\cite{huang2025self} and Causal Forcing~\cite{zhu2026causal}. We set the sink size to $3$, the force-preserved recent window to $4$, and the cache capacities of Anchor, Wave, and Veil Heads to $4$, $4$, and $2$, respectively, with the Veil merge range set to $m=2$. For Tri-Pattern Head Classification, we sample 32 VBench-Long prompts and extract attention sequences from the generated 15-second videos. The sign-rate and period thresholds are set to $80\%$ and $6.4$, respectively. Customized kernels are implemented based on FlashInfer/FlashAttention backends~\cite{ye2025flashinfer,dao2022flashattention}, and all benchmarks are conducted on NVIDIA H200 GPUs.

\begin{figure*}
    \vspace{-1.0em}
     \centering
    \includegraphics[width=\textwidth]{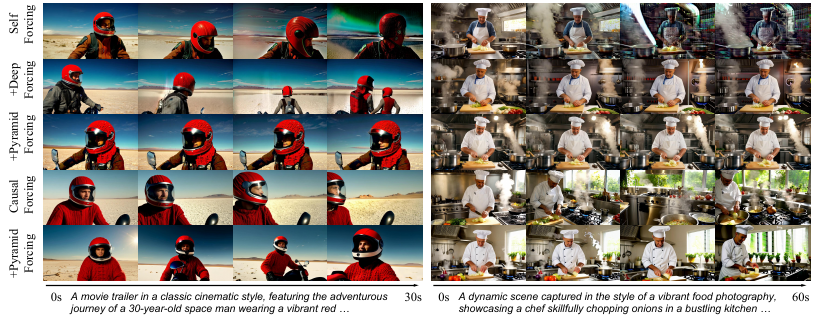}
    \captionof{figure}{
    Qualitative comparison of Pyramid Forcing and baseline methods on 30-second and 60-second video generation. Pyramid Forcing better preserves subject consistency, temporal coherence, and visual fidelity over long horizons. }
    \label{fig:6}
    \vspace{-1.0em}
\end{figure*}

\begin{table*}[t]
\centering

\begin{minipage}[t]{0.49\textwidth}
\centering
\setlength{\tabcolsep}{1pt}
\renewcommand{\arraystretch}{1.12}
\caption{Ablation study on the key components of Pyramid Forcing across six variants.}
\vspace{3pt}
\label{tab:ablation}
\resizebox{\linewidth}{!}{%
\begin{tabular}{lccccc}
\toprule
Method &
Dynamic Degree $\uparrow$ &
Image Quality $\uparrow$ &
VBench-Long $\uparrow$ &
Lat. $\downarrow$ &
Mem. $\downarrow$ \\
\midrule
Self Forcing & 37.89 & 71.51 & 80.44 & 54.12s & 35.80GB \\
+ Dynamic ROPE & 41.79 & 70.74 & 80.96 & 54.17s & 35.80GB \\
+ Ragged-Cache & 41.01 & 71.08 & 80.43 & \textbf{52.05s} & 35.80GB \\
+ Head  & 39.84 & 71.16 & 80.99 & 54.03s & \textbf{34.87GB} \\
+ Head \& Pyramid & 45.89 & \textbf{72.06} & 81.58 & 139.70s & 36.06GB \\
\rowcolor{gray!12}
Pyramid Forcing & \textbf{50.58} & 71.73 & \textbf{82.16} & 52.31s & 36.32GB \\
\bottomrule
\end{tabular}%
}
\end{minipage}
\hfill
\begin{minipage}[t]{0.49\textwidth}
\centering
\setlength{\tabcolsep}{7pt}
\renewcommand{\arraystretch}{1.12}
\caption{Efficiency and peak GPU memory usage. FPS denotes generated frames per second.}
\vspace{3pt}
\label{tab:efficiency}
\resizebox{\linewidth}{!}{%
\begin{tabular}{lcccc}
\toprule
Method &  VBench-Long $\uparrow$ & Latency $\downarrow$ & FPS $\uparrow$ & Memory $\downarrow$ \\
\midrule
Self Forcing
& 80.57 & 54.01s & 11.85 & 35.80GB \\
+ Deep Forcing
& 80.59 & 52.85s & 12.19 & 37.17GB \\
\rowcolor{gray!12}
+ Pyramid Forcing
& 82.25 & 52.18s & 11.98 & 36.32GB \\
Causal Forcing
& 79.78 & 46.75s & 13.31 & 47.93GB \\
\rowcolor{gray!12}
+ Pyramid Forcing
& 80.40 & 45.21s & 13.78 & 45.59GB \\
\bottomrule
\end{tabular}%
}
\end{minipage}

\end{table*}

\subsection{Main Results}

\paragraph{Quantitative Results.}
We evaluate 30-second and 60-second autoregressive video generation on VBench-Long, with results reported in Tab.~\ref{tab:main_results}. Pyramid Forcing achieves the highest Total Score at both lengths, improving Self Forcing from 80.57 to 82.25 on 30-second videos and from 77.87 to 81.21 on 60-second videos. It also consistently improves over Deep Forcing and further boosts Causal Forcing, demonstrating its effectiveness in mitigating long-horizon degradation. Beyond overall quality, Pyramid Forcing shows clear gains in Dynamic Degree, especially under Causal Forcing, increasing it from 61.38 to 82.67 on 30-second videos and from 57.03 to 86.39 on 60-second videos. We attribute these gains to head-aware cache allocation: unlike unified cache policies, Pyramid Forcing assigns differentiated caches according to head-specific temporal dependencies, preserving long-range consistency while avoiding overly strong motion constraints.

\paragraph{Qualitative Results.}
The visual comparisons in Fig.~\ref{fig:6} show that existing methods suffer from varying degrees of long-term degradation during long video generation. Self Forcing produces noticeable appearance drift and noise artifacts as generation progresses. Deep Forcing and Rolling Forcing can mitigate part of the degradation, but their subject details and motion dynamics remain unstable. In contrast, Pyramid Forcing better preserves subject appearance, scene structure, and local details in 60-second videos, while retaining richer camera and subject motion. These qualitative results further verify that our method achieves a better balance among long-range consistency, visual quality, and dynamic expressiveness.

\begin{wrapfigure}{l}{0.45\textwidth}
\vspace{-8pt}
\centering
\includegraphics[width=\linewidth]{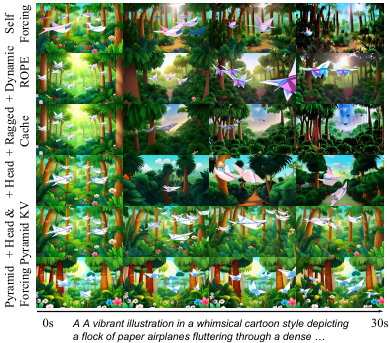}
\caption{Qualitative ablation of components.}
\label{fig:7}
\vspace{-12pt}
\end{wrapfigure}

\subsection{Ablation Studies}

\paragraph{Ablation Study on Key Components.}
We conduct a component-wise ablation study over six variants: Self Forcing, variants with only Dynamic RoPE, Ragged-Cache Attention, or Head Classification, the combination of Head Classification and Pyramid KVCache policies, and the full Pyramid Forcing. For the Head Classification variant, the type-specific neighboring windows are set to $\mathcal{R}^{(A)}=12$, $\mathcal{R}^{(W)}=10$, and $\mathcal{R}^{(V)}=5$.
    As shown in Tab.~\ref{tab:ablation} and Fig.~\ref{fig:7}, Dynamic RoPE improves motion dynamics, and Ragged-Cache Attention reduces latency with negligible quality loss. Head Classification improves VBench-Long, while Pyramid KVCache policies further enhance quality at the cost of extra heterogeneous-cache overhead. The full Pyramid Forcing achieves the best Dynamic Degree and VBench-Long with low latency, confirming the complementarity of all components.

\paragraph{Efficiency and Peak GPU Memory Usage Study.}
We evaluate the computational efficiency and peak GPU memory usage of Pyramid Forcing against other KVCache strategies, with results reported in Tab.~\ref{tab:efficiency}. Pyramid Forcing achieves comparable efficiency on both Self Forcing and Causal Forcing while consistently improving VBench-Long performance.

\paragraph{Effect of Prompt Number and Video Duration on Offline Tri-pattern Head Classification.}
We study how prompt number and video duration affect classification quality and overhead on a single H200 GPU. As shown in Tab.~\ref{tab:classification}, 32 prompts with 15-second videos achieve comparable quality to longer or larger settings while substantially reducing time and memory cost. Since classification is performed only once per model, its overhead is negligible compared with training or distillation.

\paragraph{Effect of head-specific policy parameters in Pyramid KVCache policies.}
We evaluate the sensitivity of Pyramid KVCache policies to head-specific parameters. As shown in Tab.~\ref{tab:head_policy}, all configurations outperform Self Forcing and remain close to the full Pyramid Forcing performance. The small variation across settings indicates that the method is robust to hyperparameter choices.


\begin{table*}[t]
\centering

\begin{minipage}[t]{0.49\textwidth}
\vspace{0pt}
\centering
\setlength{\tabcolsep}{5pt}
\renewcommand{\arraystretch}{1.12}
\caption{Effect of prompt number and video duration on offline tri-pattern head classification.}
\vspace{3pt}
\label{tab:classification}
\resizebox{\linewidth}{!}{%
\begin{tabular}{llccc}
\toprule
Prompts & Duration & VBench-Long $\uparrow$ & Overhead $\downarrow$ & Peak Mem. $\downarrow$ \\
\midrule
1   & 60s & 81.83 & 507.32 s & 80.14 GB \\
1   & 30s & 81.81 & 249.58 s & 62.20 GB \\
1   & 15s & 81.83 & 103.26 s & 38.81 GB \\
1   & 10s & 81.60 & 70.10 s & 32.75 GB \\
1   & 5s  & 81.51 & 33.95 s & 26.11 GB \\
\midrule
128 & 15s & 81.83 & 3.67 h & 38.81 GB \\
64  & 15s & 81.83 & 1.83 h & 38.81 GB \\
32  & 15s & 81.83 & 0.91 h & 38.79 GB \\
16  & 15s & 81.82 & 0.46 h & 38.80 GB \\
\bottomrule
\end{tabular}%
}
\end{minipage}
\hfill
\begin{minipage}[t]{0.49\textwidth}
\vspace{0pt}
\centering
\setlength{\tabcolsep}{3pt}
\renewcommand{\arraystretch}{1.12}
\caption{Effect of head-specific policy parameters in Pyramid KVCache policies.}
\vspace{3pt}
\label{tab:head_policy}
\resizebox{\linewidth}{!}{%
\begin{tabular}{llccc}
\toprule
Head Type & Setup & Dynamic Degree $\uparrow$ & Image Quality $\uparrow$ & VBench-Long $\uparrow$ \\
\midrule
Self Forcing & - & 37.89 & 71.51 & 80.44 \\
Pyramid Forcing & - & 50.58 & 71.73 & 82.16 \\
\midrule
\multirow{3}{*}{\shortstack{Anchor\\Head}}
& $s_t=3$  & 50.98 & 72.24 & 81.92 \\
& $s_t=6$  & 51.76 & 72.10 & 82.05 \\
& $s_t=12$ & 53.32 & 72.80 & 82.08 \\
\midrule
\multirow{3}{*}{\shortstack{Wave\\Head}}
& $P=3 $ & 48.05 & 72.40 & 81.95 \\
& $P=6$  & 53.32 & 72.49 & 82.06 \\
& $P=12$& 50.78 & 72.01 & 81.96 \\
\midrule
\multirow{2}{*}{\shortstack{Veil\\Head}}
& $m=2$ & 50.39 & 72.23 & 82.00 \\
& $m=3$ & 48.63 & 72.23 & 81.94 \\
\bottomrule
\end{tabular}%

}
\end{minipage}

\end{table*}
\section{Related Work}

\paragraph{Autoregressive Video Diffusion.}
Diffusion-based video generators achieve strong visual quality through large-scale training~\citep{peebles2023scalable,wan2025wan,yang2024cogvideox}, but full-sequence denoising with bidirectional attention limits them to fixed temporal windows. Recent autoregressive video diffusion methods extend generation by conditioning future clips on synthesized history: CausVid~\citep{yin2025slow} distills causal generators from bidirectional DiTs, Self Forcing~\citep{huang2025self} reduces train-test mismatch via self-generated conditioning, and Causal Forcing~\citep{zhu2026causal} improves distillation with ODE initialization. Self Forcing++~\citep{cui2025self}, Rolling Forcing~\citep{liu2025rolling}, and LongLive~\citep{yang2025longlive} further improve long-horizon stability through error correction or training-inference alignment. Nevertheless, long video generation still suffers from appearance drift, motion decay, and identity inconsistency.

\paragraph{KVCache Policy.} KVCache optimization is crucial for improving the quality and efficiency of autoregressive generation. Existing LLM studies mainly focus on cache selection~\cite{li2024snapkv,zhang2023h2o,xiao2023efficient,lee2024infinigen}, budget allocation~\cite{cai2024pyramidkv,feng2024ada,yang2024pyramidinfer}, and cache compression~\cite{cheng2025qaq,kim2023compressed,liu2024kivi,liu2024minicache,dong2024get}. However, these methods are tailored to textual sequences and are difficult to directly transfer to autoregressive video generation, where KVCaches must preserve semantic context, temporal continuity, spatial structure, and cross-frame dependencies. Moreover, attention heads often exhibit heterogeneous temporal behaviors, requiring video-specific cache policies.

\section{Conclusion}
\label{sec:conclusion}

We propose Pyramid Forcing, a training-free head-aware KVCache framework for autoregressive long video generation. By identifying three temporal dependency patterns of attention heads, i.e., Anchor, Wave, and Veil Heads, Pyramid Forcing applies differentiated cache policies with Ragged-Cache Attention. Experiments show improved subject consistency, temporal coherence, and visual fidelity while preserving inference efficiency, providing a new direction for fine-grained cache optimization in long-horizon video generation.

\clearpage
{
\small
\bibliographystyle{unsrt}
\bibliography{_ref/1_AR_DiT}
}






\clearpage
\appendix

\section{Additional Visualizations and Experimental Results.}
\label{sec:appendix_results}

\FloatBarrier
\subsection{Additional Evaluation Metrics of the Main Experiment}
\label{sec:appendix_main}

Table~\ref{tab:supp_results} reports the remaining VBench-Long metrics for the main comparison, complementing the qualitative examples above and the ablations in Section~\ref{sec:sink_recent_impact}.

\begin{table*}[!htbp]
\small
\centering
\setlength{\tabcolsep}{2.5pt} 
\renewcommand{\arraystretch}{1.15}
\caption{Supplementary quantitative comparison on long video generation. This table includes the remaining 8 VBench-Long metrics not shown in the main text.}
\vspace{4pt}
\label{tab:supp_results}
\resizebox{\textwidth}{!}{%
\begin{tabular}{lcccccccc}
\toprule
\multicolumn{1}{c}{\multirow{2}{*}{Model}} &
App. & Background & Human & \multirow{2}{*}{Scene $\uparrow$} & Subject & Temporal & Temporal & Temporal \\
& Style $\uparrow$ & Consist. $\uparrow$ & Action $\uparrow$ & & Consist. $\uparrow$ & Flickering $\downarrow$ & Style $\uparrow$ & (2D) $\uparrow$ \\
\midrule
& \multicolumn{8}{c}{\textit{30 seconds}} \\
\cmidrule{2-9}
CausVid             & 24.19 & 97.04 & 18.70 & 26.76 & 98.05 & 97.12 & 22.39 & 56.36 \\
Rolling Forcing     & 23.91 & 96.66 & 20.02 & \textbf{34.77} & \textbf{97.71} & 97.69 & 24.85 & 56.93 \\
LongLive            & 23.61 & 96.62 & 21.14 & 31.50 & 97.61 & 97.69 & 24.54 & 57.33 \\
\cmidrule{2-9}
Self Forcing        & 23.89 & 95.84 & 21.14 & 30.90 & 96.31 & 97.18 & 24.18 & 57.14 \\
+ Deep Forcing      & \textbf{24.08} & 95.78 & \textbf{21.44} & 30.62 & 96.19 & \textbf{96.86} & 24.35 & 57.66 \\
\rowcolor{gray!12}
+ Pyramid Forcing   & 23.82 & \textbf{96.30} & 20.75 & 32.61 & \textbf{97.28} & 97.35 & \textbf{24.55} & \textbf{58.41} \\
\cmidrule{2-9}
Causal Forcing      & 23.39 & \textbf{95.27} & 19.78 & \textbf{34.36} & \textbf{96.62} & 95.45 & 23.03 & 58.21 \\
\rowcolor{gray!12}
+ Pyramid Forcing   & \textbf{23.89} & 94.88 & \textbf{23.83} & 33.23 & 95.25 & \textbf{94.50} & \textbf{23.21} & \textbf{60.05} \\
\midrule
& \multicolumn{8}{c}{\textit{60 seconds}} \\
\cmidrule{2-9}
CausVid             & 24.16 & 97.18 & 18.80 & 25.43 & 98.18 & 97.24 & 21.65 & 55.93 \\
Rolling Forcing     & 23.89 & 96.64 & 20.49 & 32.51 & 97.74 & 97.72 & 24.20 & 56.65 \\
LongLive            & 23.80 & 96.49 & 20.82 & \textbf{33.64} & 97.44 & 97.68 & \textbf{24.54} & 57.43 \\
\cmidrule{2-9}
Self Forcing        & 23.56 & 95.53 & 18.22 & 23.95 & 95.75 & \textbf{96.85} & 22.33 & 55.00 \\
+ Deep Forcing      & \textbf{24.15} & 96.15 & 21.09 & \textbf{33.24} & 96.93 & 97.41 & \textbf{24.48} & 57.36 \\
\rowcolor{gray!12}
+ Pyramid Forcing   & 23.94 & \textbf{96.20} & \textbf{21.62} & 29.90 & \textbf{97.30} & 97.30 & 24.16 & \textbf{58.09} \\
\cmidrule{2-9}
Causal Forcing      & 23.33 & \textbf{95.09} & 19.63 & 33.20 & \textbf{96.57} & 95.39 & \textbf{22.51} & 57.44 \\
\rowcolor{gray!12}
+ Pyramid Forcing   & \textbf{23.69} & 94.58 & \textbf{22.96} & \textbf{33.44} & 94.78 & \textbf{94.00} & 22.41 & \textbf{59.91} \\
\bottomrule
\end{tabular}%
}
\end{table*}

\FloatBarrier

\subsection{Further Visualizations of the Main Experiment}

\begin{figure}[!htbp]
    \vspace{-6.0em}
    \centering
    \includegraphics[width=0.85\textwidth]{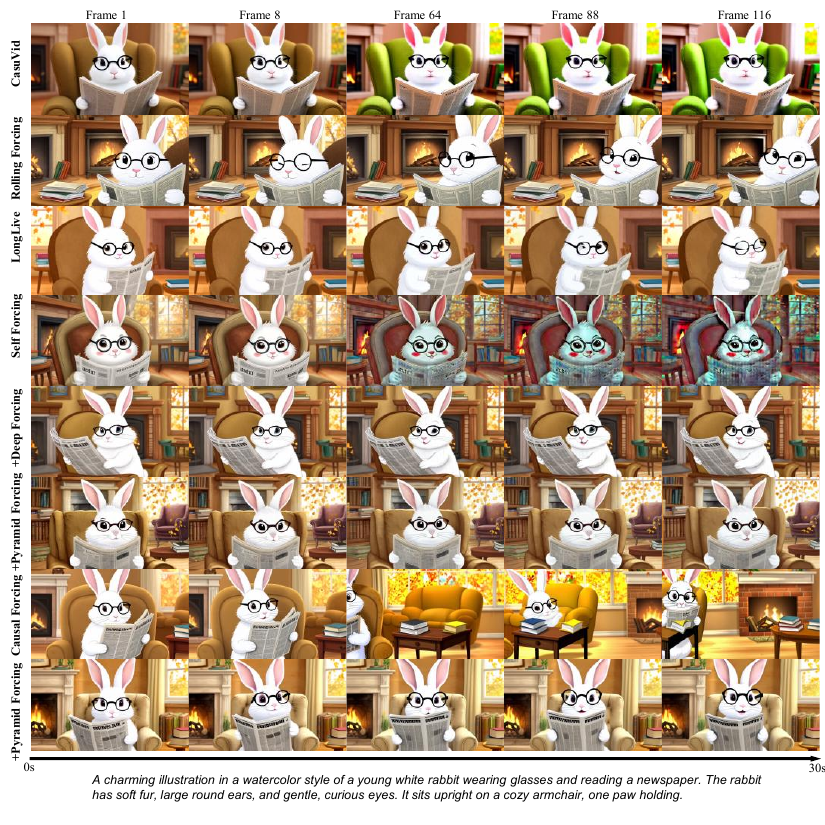}
    \vspace{-1.0em}
    \caption{
    Additional Visualization A
    }
    \label{fig:vis_a}
\end{figure}

\begin{figure}[!htbp]
    \vspace{-2.0em}
    \centering
    \includegraphics[width=0.85\textwidth]{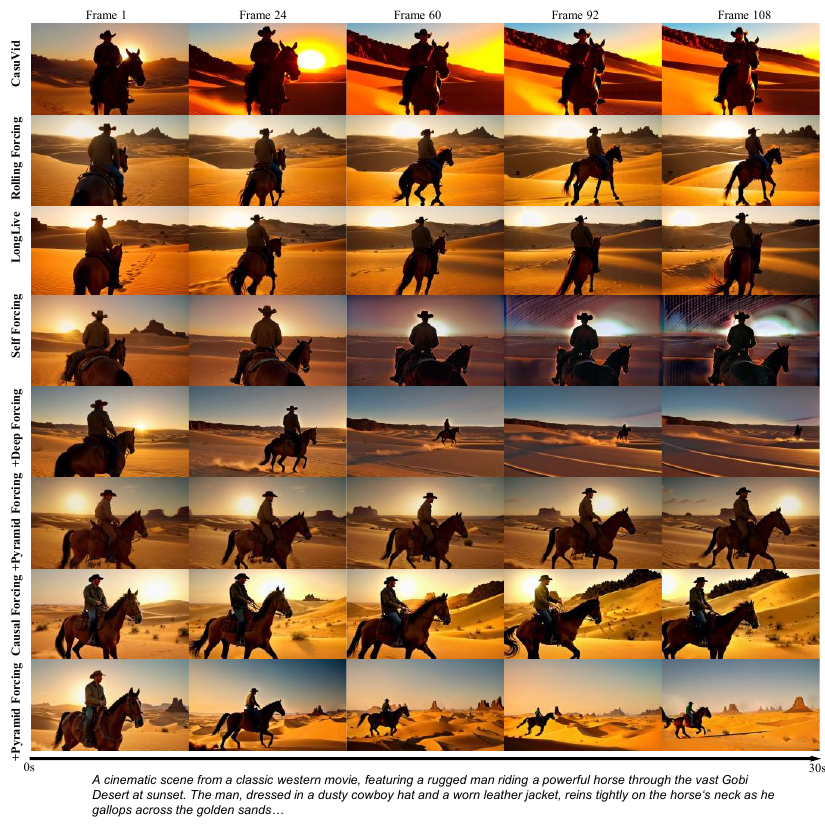}
    \vspace{-1.0em}
    \caption{
    Visualization B
    }
    \label{fig:vis_b}
\end{figure}

\begin{figure}[!htbp]
    \vspace{-6.0em}
    \centering
    \vspace{-1.0em}
    \includegraphics[width=0.85\textwidth]{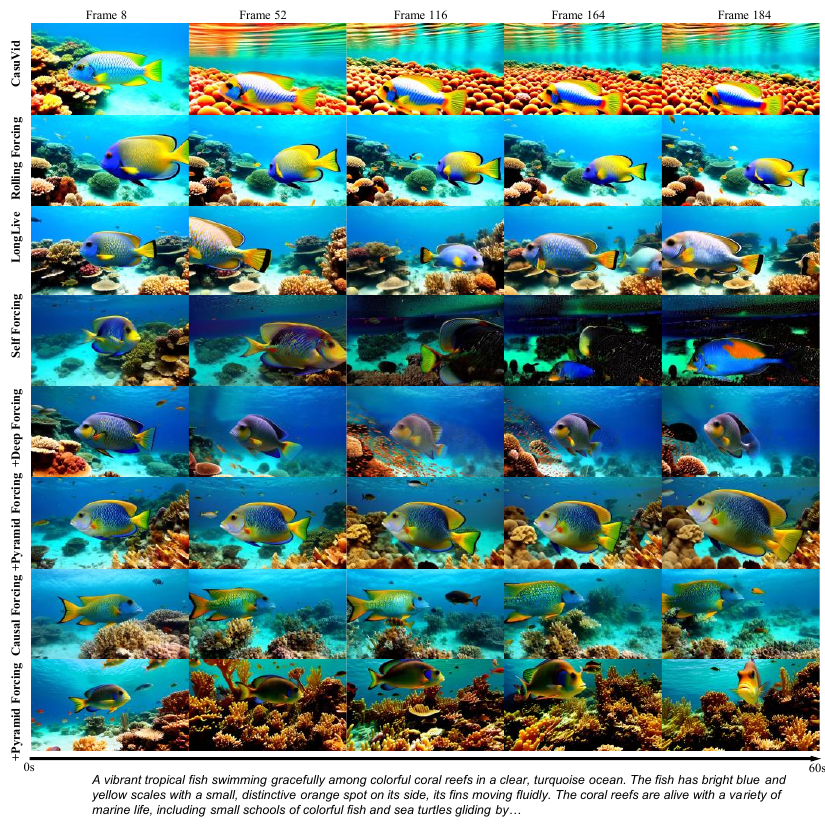}
    \caption{
    Visualization C
    }
    \label{fig:vis_c}
\end{figure}

\begin{figure}[!htbp]
    \vspace{-2.0em}
    \centering
    \vspace{-1.0em}
    \includegraphics[width=0.85\textwidth]{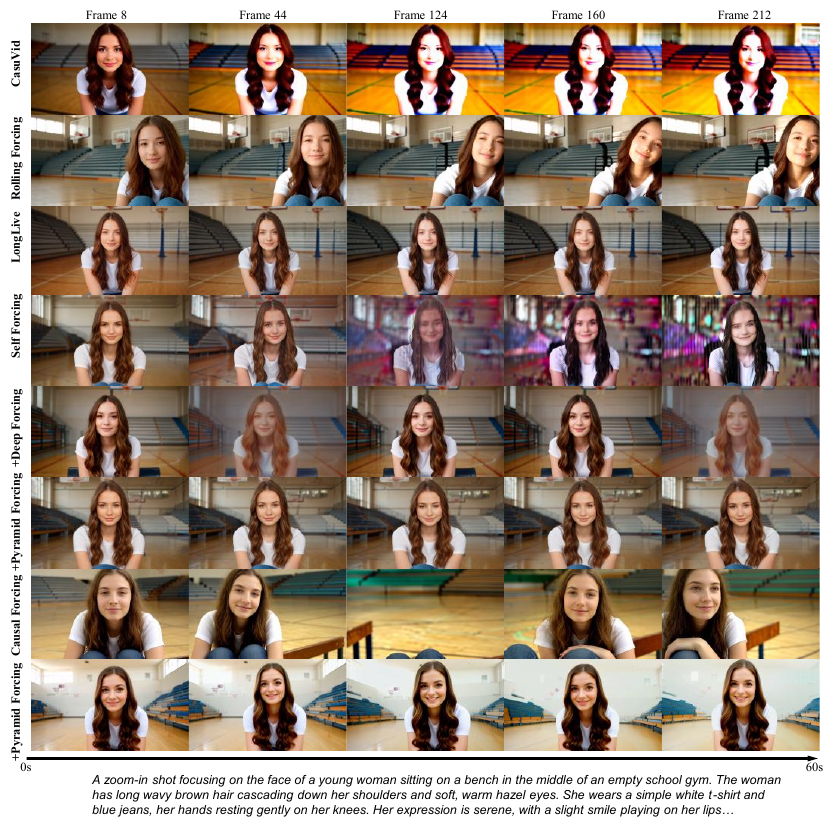}
    \caption{
    Visualization D
    }
    \label{fig:vis_d}
\end{figure}

To further facilitate visual comparison, we provide additional qualitative results in Figures~\ref{fig:vis_a}--\ref{fig:vis_d}, covering both 30-second and 60-second generation. These examples complement the main qualitative comparison by showing object-centric scenes, large camera/object motion, underwater environments, and human portraits. Overall, Pyramid Forcing preserves subject identity, scene layout, and local visual details more consistently across long horizons, while several baseline cache strategies exhibit either appearance drift, abrupt background changes, or weakened temporal continuity.

\FloatBarrier


\FloatBarrier


\FloatBarrier


\FloatBarrier

Figure~\ref{fig:vis_d} further evaluates a 60-second human portrait in an indoor gymnasium, where face details, hair, pose, and background elements are all sensitive to identity drift. Pyramid Forcing better preserves the person and the indoor scene across time, whereas several baselines show more noticeable changes in facial appearance, body configuration, or background consistency. These visual results motivate the following ablation analyses, which isolate how persistent sink-frame context and recent-frame retention contribute to stable long-horizon generation.

\FloatBarrier

\subsection{Impact of Sink Frames and Forced Recent Frame Retention}
\label{sec:sink_recent_impact}
We evaluated the impact of varying numbers of sink frames ($S$) and forced recent frames ($R$) on video quality using a single NVIDIA H200 GPU. These ablations complement the qualitative results above by testing two cache design choices that directly affect long-range temporal stability: persistent sink-frame context and forced retention of the most recent frames. As shown in Table \ref{tab:ablation_r_full_metrics}, with $S=3$ fixed, $R=4$ yields the optimal generation quality. As indicated in Table \ref{tab:ablation_s_full_metrics}, when $R=4$ is fixed, although $S=3$ is not the absolute best, the gap in the total VBench score is negligible compared to the optimal configuration.

\begin{table*}[!htp]
\centering
\caption{Ablation Study on Different Recent Window Size $R$ (Fixed $S=3$).}
\label{tab:ablation_r_full_metrics}
\renewcommand{\arraystretch}{1.2}
\setlength{\tabcolsep}{3pt}
\small
\resizebox{\textwidth}{!}{%
\begin{tabular}{lcccccccc}
\toprule
\textbf{Recent} & \textbf{Dynamic $\uparrow$} & \textbf{Motion $\uparrow$} & \textbf{Overall $\uparrow$} & \textbf{Imaging $\uparrow$} & \textbf{Aesthetic $\uparrow$} & \textbf{Quality $\uparrow$} & \textbf{Semantic $\uparrow$} & \textbf{Total $\uparrow$} \\
\textbf{Frames} & \textbf{Degree} & \textbf{Smooth.} & \textbf{Consist.} & \textbf{Quality} & \textbf{Quality} & \textbf{Score} & \textbf{Score} & \textbf{Score} \\
\midrule
2 & 19.17 & \textbf{99.09} & 25.97 & \textbf{73.01} & 65.40 & 86.79 & 54.34 & 80.30 \\
3 & 40.00 & 98.73 & \textbf{26.18} & 72.79 & \textbf{65.62} & 87.68 & \textbf{55.22} & 81.19 \\
4 & \textbf{55.63} & 98.78 & 25.52 & 72.27 & 64.58 & \textbf{88.61} & 54.30 & \textbf{81.75} \\
5 & 52.71 & 98.81 & 25.68 & 72.86 & 64.86 & 88.51 & 54.38 & 81.68 \\
6 & 53.96 & 98.79 & 25.48 & 72.79 & 64.42 & 88.48 & 54.38 & 81.66 \\
\bottomrule
\end{tabular}%
}
\end{table*}

\begin{table*}[!htbp]
\centering
\caption{Ablation Study on Different Sink Frames $S$ (Fixed $R=4$).}
\label{tab:ablation_s_full_metrics}
\renewcommand{\arraystretch}{1.2}
\setlength{\tabcolsep}{3pt}
\small
\resizebox{\textwidth}{!}{%
\begin{tabular}{lcccccccc}
\toprule
\textbf{Sink} & \textbf{Dynamic $\uparrow$} & \textbf{Motion $\uparrow$} & \textbf{Overall $\uparrow$} & \textbf{Imaging $\uparrow$} & \textbf{Aesthetic $\uparrow$} & \textbf{Quality $\uparrow$} & \textbf{Semantic $\uparrow$} & \textbf{Total $\uparrow$} \\
\textbf{Frames} & \textbf{Degree} & \textbf{Smooth.} & \textbf{Consist.} & \textbf{Quality} & \textbf{Quality} & \textbf{Score} & \textbf{Score} & \textbf{Score} \\
\midrule
1 & \textbf{67.29} & 98.64 & 25.27 & 71.38 & 62.93 & \textbf{88.77} & 53.35 & 81.69 \\
2 & 58.54 & 98.77 & \textbf{25.81} & 72.18 & 64.28 & 88.68 & \textbf{54.91} & \textbf{81.93} \\
3 & 55.63 & 98.78 & 25.52 & 72.27 & 64.58 & 88.61 & 54.30 & 81.75 \\
4 & 56.46 & 98.79 & 25.52 & \textbf{72.27} & \textbf{64.59} & 88.66 & 54.15 & 81.76 \\
5 & 57.08 & \textbf{98.79} & 25.52 & 72.27 & 64.59 & 88.72 & 54.26 & 81.83 \\
\bottomrule
\end{tabular}%
}
\end{table*}

\FloatBarrier

\subsection{Ablation Study on Period and Sign Rate Thresholds}

As shown in Table \ref{tab:ablation_period_threshold}, we conducted a comprehensive ablation study on the \textbf{Period Threshold} within our attention head classification method. The results indicate that with a fixed sign rate threshold of 80\%, the model achieves its peak performance of 82.16 when the period threshold is set to 6.4. Simultaneously, as illustrated in Table \ref{tab:ablation_symbol_rate_threshold}, we analyzed the impact of the \textbf{Sign Rate Threshold}. The experimental data demonstrate that, under a fixed period threshold of 6.4, a sign rate threshold of 80\% yields the optimal overall score for the generated videos. These findings suggest that this specific parameter configuration provides the superior classification of attention heads, establishing an ideal balance for distinguishing between Anchor/Veil heads and Wave heads.

\begin{table*}[!htbp]
\centering
\caption{Ablation Study on Period Threshold (Fixed Symbol Rate Threshold = 80\%).}
\label{tab:ablation_period_threshold}
\renewcommand{\arraystretch}{1.2}
\setlength{\tabcolsep}{3pt}
\small
\resizebox{\textwidth}{!}{%
\begin{tabular}{lcccccccc}
\toprule
\textbf{Period} & \textbf{Dynamic $\uparrow$} & \textbf{Motion $\uparrow$} & \textbf{Overall $\uparrow$} & \textbf{Imaging $\uparrow$} & \textbf{Aesthetic $\uparrow$} & \textbf{Quality $\uparrow$} & \textbf{Semantic $\uparrow$} & \textbf{Total $\uparrow$} \\
\textbf{Threshold} & \textbf{Degree} & \textbf{Smooth.} & \textbf{Consist.} & \textbf{Quality} & \textbf{Quality} & \textbf{Score} & \textbf{Score} & \textbf{Score} \\
\midrule
6.4 & \textbf{50.59} & 98.72 & 26.00 & 71.74 & 65.23 & \textbf{88.32} & \textbf{57.54} & \textbf{82.16} \\
6.5 & 46.48 & 98.78 & 26.23 & 72.72 & \textbf{65.43} & 88.07 & 56.93 & 81.85 \\
7.0 & 46.09 & \textbf{98.80} & 26.14 & \textbf{72.74} & 65.50 & 88.11 & 56.26 & 81.74 \\
8.0 & 46.29 & 98.78 & \textbf{26.27} & 72.64 & 65.39 & 88.06 & 56.12 & 81.67 \\
9.0 & 46.29 & 98.79 & 26.19 & 72.77 & 65.30 & 88.08 & 56.61 & 81.79 \\
\bottomrule
\end{tabular}%
}
\end{table*}

\begin{table*}[!htbp]
\centering
\caption{Ablation Study on Symbol Rate Threshold (Fixed Period Threshold = 6.4).}
\label{tab:ablation_symbol_rate_threshold}
\renewcommand{\arraystretch}{1.2}
\setlength{\tabcolsep}{3pt}
\small
\resizebox{\textwidth}{!}{%
\begin{tabular}{lcccccccc}
\toprule
\textbf{Symbol Rate} & \textbf{Dynamic $\uparrow$} & \textbf{Motion $\uparrow$} & \textbf{Overall $\uparrow$} & \textbf{Imaging $\uparrow$} & \textbf{Aesthetic $\uparrow$} & \textbf{Quality $\uparrow$} & \textbf{Semantic $\uparrow$} & \textbf{Total $\uparrow$} \\
\textbf{Threshold} & \textbf{Degree} & \textbf{Smooth.} & \textbf{Consist.} & \textbf{Quality} & \textbf{Quality} & \textbf{Score} & \textbf{Score} & \textbf{Score} \\
\midrule
70\% & 45.31 & 98.74 & 25.90 & \textbf{72.78} & 65.12 & 87.86 & 56.32 & 81.55 \\
80\% & \textbf{50.59} & 98.72 & \textbf{26.00} & 71.74 & \textbf{65.23} & \textbf{88.32} & \textbf{57.54} & \textbf{82.16} \\
90\% & 49.22 & \textbf{98.84} & 25.72 & 71.34 & 64.38 & 88.09 & 56.84 & 81.84 \\
\bottomrule
\end{tabular}%
}
\end{table*}

\FloatBarrier

\subsection{Ablation Study on Head-Specific Capacities}

As shown in Table \ref{tab:head_ablation_results}, we conducted an ablation study on the \textbf{Capacity} parameters for the three distinct types of attention heads: Anchor, Wave, and Veil.

\begin{itemize}
    \item \textbf{Anchor Head (Stride):} Anchor heads benefit from larger capacity, achieving the best performance at capacity 4, consistent with their long-range dependency requirement.
    \item \textbf{Wave Head (Periodic):} Wave heads also peak at capacity 4, indicating that sufficient capacity helps capture periodic motion patterns.
    \item \textbf{Veil Head (Merge):} Veil heads perform best with a smaller capacity, reaching 82.242, suggesting that excessive history may introduce noise rather than improve fusion.
\end{itemize}

This group of experiments fully demonstrates that adjusting the capacity specifically for different types of attention heads can maximize the optimization of video generation quality.

\begin{table*}[!htbp]
\small
\centering
\setlength{\tabcolsep}{3pt}
\renewcommand{\arraystretch}{1.15}
\caption{Quantitative comparison based on different Head Types and Capacities. We evaluate the performance across multiple quality metrics on VBench-Long Lite.}
\vspace{4pt}
\label{tab:head_ablation_results}
\resizebox{\textwidth}{!}{%
\begin{tabular}{lcccccccc}
\toprule
\multicolumn{1}{c}{\multirow{2}{*}{Capacity}} &
Dynamic & Motion & Overall & Imaging & Aesthetic & Quality & Semantic & Total \\
& Degree $\uparrow$ & Smoothness $\uparrow$ & Consistency $\uparrow$ & Quality $\uparrow$ & Quality $\uparrow$ & Score $\uparrow$ & Score $\uparrow$ & Score $\uparrow$ \\
\midrule
& \multicolumn{8}{c}{\textit{Anchor Head (Stride)}} \\
\cmidrule{2-9}
1 & 51.953 & 98.694 & 25.984 & 71.430 & \textbf{65.502} & 88.378 & \textbf{57.076} & 82.118 \\
2 & 52.148 & 98.729 & 26.042 & \textbf{71.451} & 65.352 & 88.386 & 57.017 & 82.112 \\
\rowcolor{gray!12}
4 & \textbf{52.734} & \textbf{98.751} & \textbf{26.069} & 71.429 & 65.267 & \textbf{88.482} & 56.894 & \textbf{82.165} \\
\midrule
& \multicolumn{8}{c}{\textit{Wave Head (Periodic)}} \\
\cmidrule{2-9}
3 & 50.000 & \textbf{98.756} & \textbf{26.057} & 71.512 & 65.457 & 88.287 & \textbf{57.290} & 82.088 \\
\rowcolor{gray!12}
4 & \textbf{53.125} & 98.748 & 26.024 & 71.629 & \textbf{65.496} & \textbf{88.570} & 56.718 & \textbf{82.200} \\
5 & 52.344 & 98.754 & 25.959 & \textbf{71.708} & 65.223 & 88.409 & 56.775 & 82.082 \\
\midrule
& \multicolumn{8}{c}{\textit{Veil Head (Merge)}} \\
\cmidrule{2-9}
\rowcolor{gray!12}
2 & \textbf{53.711} & 98.716 & \textbf{26.106} & 71.466 & 65.512 & \textbf{88.520} & \textbf{57.131} & \textbf{82.242} \\
3 & 52.734 & 98.751 & 25.962 & \textbf{71.719} & 65.428 & 88.492 & 56.540 & 82.102 \\
4 & 50.195 & \textbf{98.786} & 26.035 & 71.490 & \textbf{65.610} & 88.409 & 56.592 & 82.046 \\
\bottomrule
\end{tabular}%
}
\end{table*}

\FloatBarrier

\vspace{-4.0em}
\subsection{Visualizing Attention Patterns}
\begin{figure}[!htbp]
    \centering
    \includegraphics[width=0.96\textwidth]{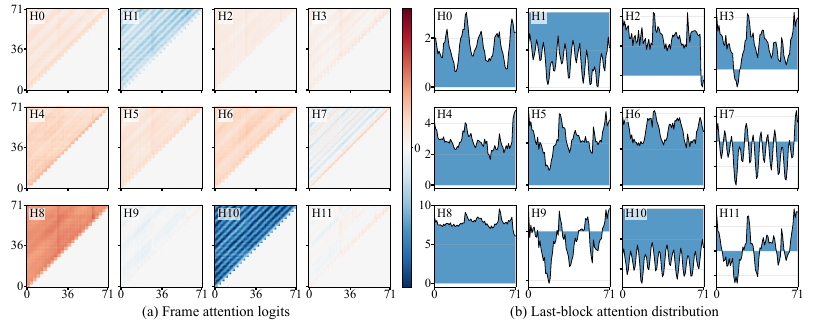}
    \caption{
    Attention visualization of the 72-frame Self Forcing model at Layer 23: ``A FPV drone shot through a castle on a cliff.''
    }
    \label{fig:selfforcing_castle72_layer23}
\end{figure}

\begin{figure}[!htbp]
    \centering
    \includegraphics[width=0.96\textwidth]{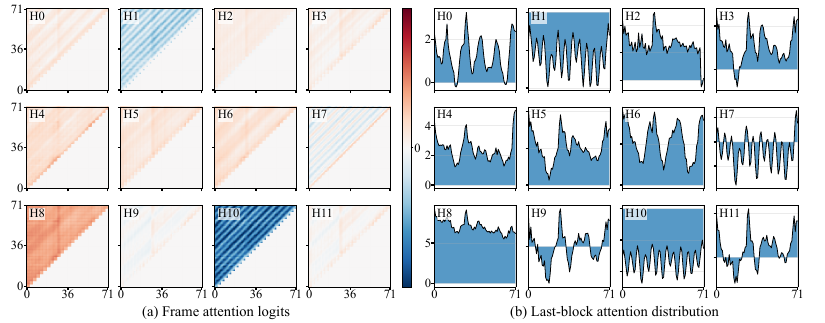}
    \caption{
    Attention visualization of the 72-frame Self Forcing model at Layer 23: ``Super fast zoom out from the peak of a frozen mountain where a lonely hiker is arriving to the summit.''
    }
    \label{fig:selfforcing_hiker72_layer23}
\end{figure}

\begin{figure}[!htbp]
    \centering
    \includegraphics[width=0.96\textwidth]{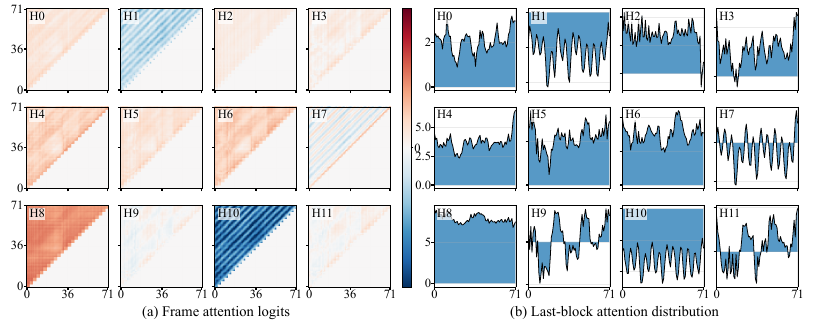}
    \caption{
    Attention visualization of the 72-frame Self Forcing model at Layer 23: ``A white and orange tabby cat is seen happily darting through a dense garden, cinematic warm tones, shallow depth of field.''
    }
    \label{fig:selfforcing_cat72_layer23}
\end{figure}

\begin{figure}[!htbp]
    \centering
    \includegraphics[width=0.96\textwidth]{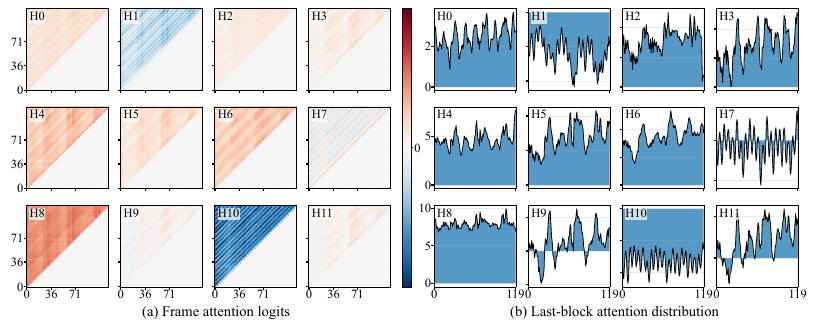}
    \caption{
    Attention visualization of the 120-frame Self Forcing model at Layer 23: ``A majestic eagle soaring through a cloudy sky, cinematic lighting.''
    }
    \label{fig:selfforcing_eagle120_layer23}
\end{figure}

Attention heads exhibit stable, repeatable functional patterns across diverse prompts and lengths. As shown in Figures~\ref{fig:selfforcing_castle72_layer23}--\ref{fig:selfforcing_eagle120_layer23}, despite varying scene semantics and dynamics, specific head groups consistently demonstrate long-range, oscillatory, or local attention. This suggests that functional specialization is an intrinsic structural property of autoregressive video models.

\begin{figure}[!htbp]
    \centering
    \includegraphics[width=0.96\textwidth]{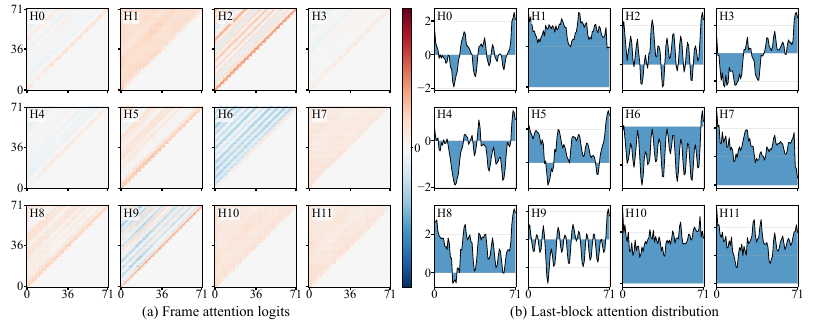}
    \caption{
    Attention visualization of the 72-frame Causal Forcing model at Layer 23: ``A majestic eagle soaring through a cloudy sky, cinematic lighting.''
    }
    \label{fig:causalforcing_eagle72_layer23}
\end{figure}

Moreover, this phenomenon is not limited to Self Forcing. As illustrated in Figure~\ref{fig:causalforcing_eagle72_layer23}, Causal Forcing also demonstrates clear functional differentiation among attention heads. Similar to Self Forcing, the attention heads can likewise be broadly categorized into these three functional types.

\begin{figure}[!htbp]
    \centering
    \includegraphics[width=0.96\textwidth]{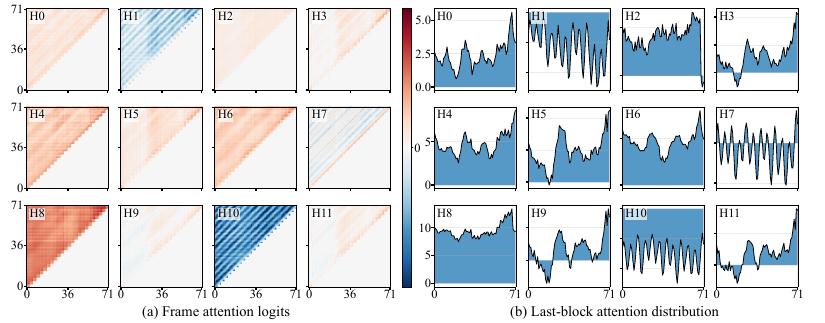}
    \caption{
    Attention visualization of the 72-frame Self Forcing model at Layer 23: ``A majestic eagle soaring through a cloudy sky, cinematic lighting.''
    }
    \label{fig:selfforcing_eagle72_layer23}
\end{figure}

\vspace{-2.0em}

\vspace{-2.0em}


\FloatBarrier

\subsection{Visualization of Classification Results}

\begin{figure}[!htbp]
    \centering
    \includegraphics[width=\textwidth]{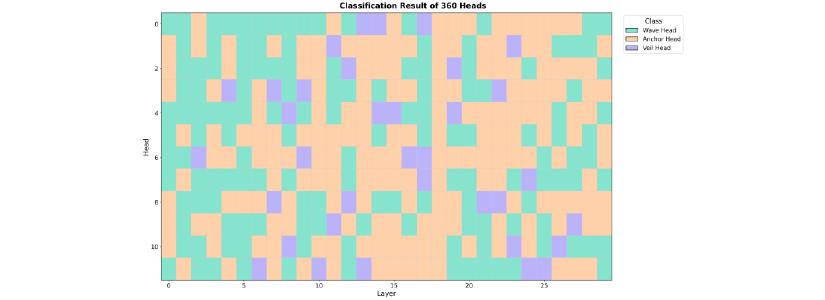}
    \caption{
    Aggregated classification results obtained via majority voting across 256 prompts (15s video duration), using a period threshold of 6.4 and a sign rate threshold of 80\%.
    }
    \label{fig:classification_results}
\end{figure}
\FloatBarrier

\subsection{Periodicity Distribution across Additional Layers}

During the FFT analysis stage, we process attention sequences for each of the 256 prompts across 30 layers and 12 heads per layer, using the frame range from 0 to 68. For each sequence, we first apply a first-order difference to emphasize inter-frame variation, followed by mean removal and a Hanning window. We then compute the spectrum with \texttt{rfft}, discard the zero-frequency component, and convert frequencies into periods as $\mathrm{period}=1/f$. The spectral peak is not selected solely by raw amplitude. Instead, we perform harmonic folding, where the energy from multiple harmonics is folded back to the fundamental frequency with decaying weights. The period with the highest folded score is selected as the top-1 period estimate for the corresponding head under each prompt. Finally, the 256 prompt-level estimates are aggregated to obtain summary statistics, including the mean, standard deviation, minimum/maximum values, and dominant period.

\begin{figure}[!htbp]
    \centering
    \includegraphics[width=\textwidth]{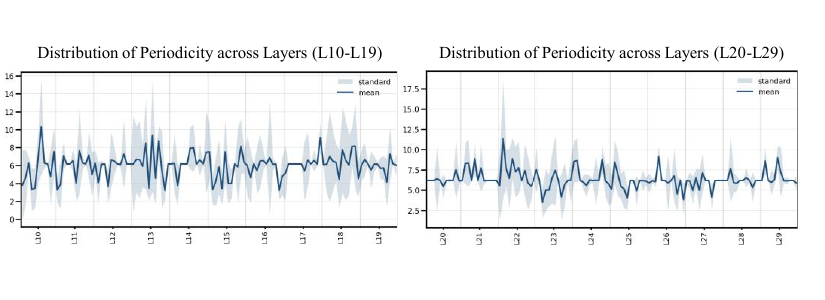}
    \caption{
    The periodicity distribution across layers L10--L29 reveals that most attention heads consistently concentrate around a dominant period of 6.0, providing a stable statistical basis for our head-aware classification.
    }
    \label{fig:periodicity_distribution}
\end{figure}
\FloatBarrier

\section{Algorithmic Analysis}

\subsection{Pre-Softmax Logits for Head Classification}

In constructing a classification system to identify attention heads, the choice of classification metrics is crucial. In the standard attention mechanism, features are determined by normalized weights:
\begin{equation}
\text{Attn Score} = \text{Softmax}(z_i) = e^{z_i} / \sum_{j} e^{z_j}
\end{equation}
However, to better enhance the classification performance, we select the \textbf{raw logits before the Softmax operation} (i.e., $z = QK^T / \sqrt{d_k}$) as the basis for attention head classification.

From a geometric representation perspective, the logit space preserves the \textbf{polarity information} of the vectors, which is absent in normalized probability distributions. In the original dot-product space, if the Query and Key vectors point in opposite directions (with an angle greater than 90$^\circ$), the corresponding logit value is negative. Physically, this represents the model's explicit inhibition of specific redundant information. 

However, the Softmax function restricts the codomain to $(0, 1)$, forcibly converting this directional ``negative correlation inhibition'' into positive probability weights. This transition from ``directional vector projection'' to ``positive probability measure'' obscures the original intent of the attention head to actively suppress invalid information. Therefore, utilizing logits for classification can more authentically restore the polarity of the model during feature extraction, enabling a more accurate functional definition of attention heads in long video generation.
\subsection{Failure Cases of Head Classification}

\begin{figure}[!htbp]
    \centering
    \begin{subfigure}[b]{0.48\textwidth}
        \centering
        \includegraphics[width=\textwidth]{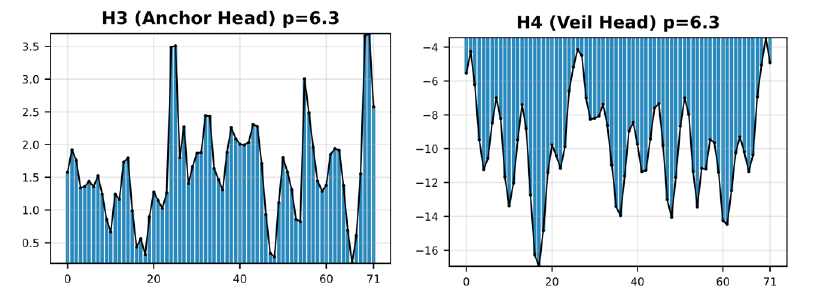} 
        \caption{Pseudo-periodic logit sequences showing one-sided bias.}
        \label{fig:pseudo_periodic}
    \end{subfigure}
    \hfill
    \begin{subfigure}[b]{0.48\textwidth}
        \centering
        \includegraphics[width=\textwidth]{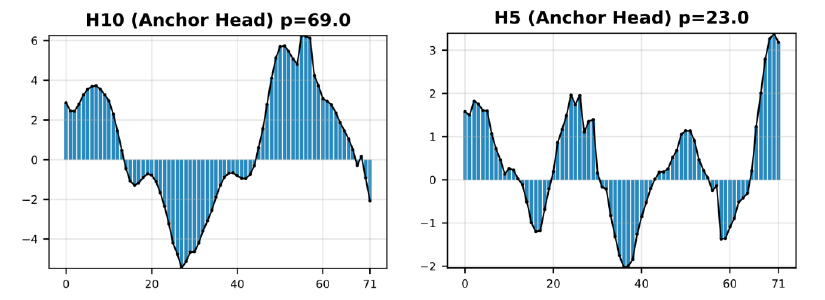} 
        \caption{Boundary samples near the classification threshold.}
        \label{fig:boundary_cases}
    \end{subfigure}
    \caption{Visual analysis of head classification failure cases. (a) shows heads with periodic peaks but unbalanced sign rates; (b) illustrates samples requiring the Mean Logit fallback mechanism.}
    \label{fig:classification_failures}
\end{figure}

Classification relying solely on the FFT dominant period has inherent limitations. Some \textit{Anchor Heads} and \textit{Veil Heads} may exhibit minor periodic peaks; however, their logit sequences do not possess stable alternating positive and negative values, manifesting instead as one-sided bias or non-stationary fluctuations, as shown in Fig. \ref{fig:pseudo_periodic}. If a simple periodicity threshold is applied, these ``pseudo-periodic'' heads are easily misclassified as \textit{Wave Heads}, subsequently impacting the KVCache strategy. To address this, we introduce the \textbf{Sign Rate} to measure the balance of positive and negative logits. Genuine \textit{Wave Heads} typically exhibit a low and relatively balanced sign rate, whereas pseudo-periodic \textit{Anchor/Veil Heads} are often dominated by a single sign, leading to a significantly higher sign rate. Therefore, the sign rate threshold serves as a pre-filtering condition to effectively exclude non-oscillatory low-period heads.

Nevertheless, a small number of heads remain near the classification boundaries for both period and sign rate, making stable determination difficult based on these two metrics alone, as shown in Fig. \ref{fig:boundary_cases}. In response, we introduce the \textbf{Mean Logit} as a fallback criterion. \textit{Anchor Heads} typically maintain strong positive attention toward historical frames, resulting in a higher average logit; \textit{Veil Heads} show weaker or even inhibitory attention toward intermediate history, leading to a lower average logit; while the alternating positive and negative nature of \textit{Wave Heads} brings their mean value closer to neutrality. Consequently, when both periodicity and sign rate are uncertain, performing fallback classification based on the mean logit can reduce misclassification of boundary samples and enhance the robustness of the classification system.

\subsection{Error Accumulation in Autoregressive Video Diffusion Models}

Autoregressive video diffusion models are typically modeled as a hybrid generative framework that integrates chain decomposition with a denoising diffusion process. Formally, given a video frame sequence $x^{1:N} = (x^1, x^2, \dots, x^N)$ of length $N$, its joint distribution can be decomposed into the product of conditional probabilities as follows:
\begin{equation}
    p(x^{1:N}) = \prod_{i=1}^N p(x^i \mid x^{<i})
\end{equation}

In the actual inference process, the generation of each frame $x^i$ is conditioned on the sequence of previously generated historical frames $x^{<i}$. However, since the model cannot access the ground-truth observations from the training phase during inference and must instead rely on its own previously produced predictions $\hat{x}^{<i}$, this inevitably introduces the problem of \textbf{error accumulation}. Under this mechanism, a significant distributional discrepancy exists between the inference and training phases: tiny numerical biases $\epsilon$ generated at each step are propagated and accumulated along the autoregressive chain, leading to non-linear amplification. As the generation sequence $T$ extends, these biases force the latent space representation to gradually deviate from the original data manifold established during pre-training, resulting in severe semantic drift and visual degradation in long-range video dependencies.

This error accumulation phenomenon is particularly pronounced in long video generation tasks. When the generated temporal window exceeds the observational scope of the model's pre-training, the noise accumulated in the KVCache interferes with the precise retrieval of the attention mechanism, thereby inducing object distortion or visual collapse.
\subsection{Theoretical Analysis of Wave-Head Periodicity}
\label{ap:wave}

To explain why Wave Heads exhibit a stable periodicity of approximately $6$, we analyze the dominant Rotary Positional Embedding (RoPE) mechanism. In mainstream autoregressive video models, the relative position interaction between query and key is encoded via RoPE. For the $i$-th two-dimensional subspace, the attention score incorporates a term related to the relative distance $(t-s)$:
\begin{equation}
    \text{Score}(t, s) \propto \cos(\theta_i(t-s)), \quad \theta_i = 10000^{-2i/d}
\end{equation}
where $t$ and $s$ denote the current and historical frame indices. This structure inherently introduces an oscillatory component into the attention weights along the temporal dimension.

\paragraph{Derivation of the Dominant Period}
In practical models, low-frequency components ($\theta_i$ where $i$ is small) dominate the attention distribution because they provide smoother gradients and are less susceptible to noise during training. Specifically, for the first two-dimensional subspace ($i=0$), we have $\theta_0 = 1$. The corresponding oscillation term is:
\begin{equation}
    f(t, s) = \cos(t-s)
\end{equation}
The theoretical period of this continuous function is $T = 2\pi \approx 6.283$. Under discrete temporal sampling (integer frame indices), the model perceives the closest integer period:
\begin{equation}
    T_{perceived} \approx \text{round}(2\pi) = 6
\end{equation}
While RoPE contains multiple frequency components $\cos(\theta_i(t-s))$, the high-frequency terms tend to cancel out through averaging, leaving the low-frequency $\theta_0$ as the primary determinant of the attention signal.

\paragraph{Implications for KVCache}
This analysis confirms that Wave Heads are a direct manifestation of RoPE's low-frequency periodic structure in the attention space. This theoretical grounding directly supports our KVCache strategy:
\begin{itemize}
    \item \textbf{Selection:} Periodicity $\approx 6$ serves as a robust criterion for identifying Wave Heads.
    \item \textbf{Efficiency:} For Wave Heads, retaining only key frames at intervals of $T \approx 6$ captures the essential periodic features while eliminating redundant noise and reducing memory accumulation.
\end{itemize}

\section{Limitations and Discussion}
\label{sec:limit}

Pyramid Forcing designs differentiated KVCache policies for three types of attention heads, but it still relies on manually specified rules, such as cache capacity, stride, and merge range. Future work will explore online head identification, adaptive cache budget allocation, and learnable KVCache retrieval and compression strategies, enabling the cache mechanism to dynamically adapt to attention patterns and video content.



\end{document}